\documentclass{article}

\usepackage{times}
\usepackage{graphicx} 
\usepackage{caption}
\usepackage{subcaption}
\usepackage{float}

\usepackage{natbib}

\usepackage{algorithm}
\usepackage{algorithmic}

\usepackage[accepted,nohyperref]{icml2016arxiv}

\usepackage{amsmath}
\usepackage{amsfonts}
\usepackage{caption} 
\usepackage{multirow}
\usepackage{booktabs}
\usepackage{verbatim}
\usepackage{enumitem} 

\icmltitlerunning{Evasion and Hardening of Tree Ensemble Classifiers}

\newcommand{\textbfit}[1]{\textit{\textbf{#1}}}
\DeclareMathOperator*{\argmax}{arg\,max}

\begin{document} 

\twocolumn[
\icmltitle{Evasion and Hardening of Tree Ensemble Classifiers}

\icmlauthor{Alex Kantchelian}{akant@cs.berkeley.edu}
\icmlauthor{J. D. Tygar}{tygar@cs.berkeley.edu}
\icmlauthor{Anthony D. Joseph}{adj@cs.berkeley.edu}
\icmladdress{University of California, Berkeley}


\vskip 0.3in
]

\begin{abstract}
Classifier evasion consists in finding for a given instance $x$ the ``nearest'' instance $x'$ such 
that the classifier predictions of $x$ and $x'$ are different.
We present two novel algorithms for systematically computing evasions for tree ensembles such as boosted 
trees and random forests. Our first algorithm uses a Mixed Integer Linear Program solver and finds 
the \emph{optimal} evading instance under an expressive set of constraints. 
Our second algorithm trades off optimality for speed by using \emph{symbolic prediction}, a novel algorithm 
for fast finite differences on tree ensembles. On a digit recognition task, we demonstrate 
that both gradient boosted trees and random forests are extremely susceptible to evasions. Finally, 
we harden a boosted tree model without loss of predictive accuracy by augmenting the training set of 
each boosting round with 
evading instances, a technique we call \emph{adversarial boosting}.
\end{abstract} 

\section{Introduction}
Deep neural networks (DNN) represent a prominent success of machine learning.
These models can successfully and accurately address difficult learning problems, including classification of audio, video, 
and natural language possible where previous approaches have failed.
Yet, the existence of evading instances for the current incarnation of DNNs~\cite{szegedy2013} shows
a perhaps surprising brittleness: for virtually any instance $x$ that the model classifies correctly, 
it is possible to find a negligible perturbation $\delta$ such that $x + \delta$ \emph{evades} being correctly classified,
that is, receives a (sometimes widely) inaccurate prediction.

The general study of the evasion problem matters on both conceptual and practical grounds. First, we expect
a high-performance learning algorithm to generalize well and be hard to evade: only a ``large enough'' perturbation $\delta$ 
should be able to alter its decision. The existence of small-$\delta$ evading instances shows a defect in the 
generalization ability of the model, and hints at improper model class and/or insufficient regularization. 
Second, machine learning is becoming the workhorse of security-oriented applications, the most prominent example 
being unwanted content filtering. In those applications, the attacker has a large incentive for finding evading instances. 
For example, spammers look for small, cost-effective changes to their online content to avoid detection and removal. 

While prior work extensively studies the evasion problem on differentiable models by means of gradient descent, 
those results are reported in an essentially qualitative fashion, implicitly defaulting 
the choice of metric for measuring $\delta$ to the $L_2$ norm. Further, non-differentiable, non-continuous models
have received very little attention. Tree sum-ensembles as produced by boosting or bagging are perhaps the 
most important models from this class as they are often able to achieve competitive performance and enjoy 
good adoption rates in both industrial and academic contexts.

In this paper, we develop two novel exact and approximate evasion algorithms for sum-ensemble of trees. 
Our exact (or optimal) evasion algorithm computes the smallest $\delta$ according to the $L_p$ norm 
for $p=0,1,2,\infty$ such that the model misclassifies $x+\delta$. The algorithm relies on 
a Mixed Integer Linear Program solver and enables precise quantitative robustness statements. 
We benchmark the robustness of boosted trees and random forests on a concrete handwritten digit classification task 
by comparing the minimal required perturbation $\delta$ across many representative models. Those models include 
$L_1$ and $L_2$ regularized logistic regression, max-ensemble of linear classifiers (shallow max-out network), 
a 3-layer deep neural network and a classic RBF-SVM. The comparison shows that for this task, despite their 
competitive accuracies, tree ensembles are consistently the most brittle models across the board.

Finally, our approximate evasion algorithm is based on \emph{symbolic prediction}, 
a fast and novel method for computing finite differences for tree ensemble models. 
We use this method for generating more than 11 million synthetic confusing instances and incorporate those
during gradient boosting in an approach we call \emph{adversarial boosting}. This technique produces a hardened model
which is significantly harder to evade without loss of accuracy.

\section{Related Work}
From the onset of the adversarial machine learning subfield, evasion is recognized as part of the larger family of 
attacks occurring at inference time: \emph{exploratory} attacks~\cite{barreno2006}. 
While there is a prolific literature considering the evasion of linear or otherwise differentiable 
models~\cite{dalvi2004,lowd2005,lowd2005adv,nelson2012,bruckner2012,fawzi2014,biggio2013,szegedy2013,srndic2014}, 
we are only aware of a single paper tackling the case of tree ensembles. 
In Xu et al.~\cite{xuautomatically}, the authors present a genetic algorithm for finding malicious PDF instances which evade detection.

In this paper, we forgo application-specific feature extraction and directly work in feature space.
We briefly discuss 
strategies for modeling the feature extraction step in paragraph \textbf{additional constraints} 
of section~\ref{subsec:milp}.
We deliberately do not limit the amount 
of information available for carrying out evasion. In this paper, our goal is to establish the intrinsic evasion 
robustness of the machine learning models themselves, and thus provide 
a guaranteed worst-case lower-bound. In contrast to~\cite{xuautomatically}, our exact algorithm 
guarantees optimality of the solution, and our approximate algorithm performs a fast coordinate descent without the 
additional tuning and hyper-parameters that a genetic algorithm requires.

We contrast our paper with a few related papers on deep neural networks, as these are the closest in spirit to the ideas developed here. 
Goodfellow et al.~\cite{goodfellow2014} hypothesize that evasion in practical deep neural networks is possible  
because these models are locally linear. However, this paper demonstrates that despite 
their extreme non-linearity, boosted trees are even more susceptible to evasion than neural networks. 
On the hardening side, Goodfellow et al.~\cite{goodfellow2014} introduce a regularization penalty term which simulates the 
presence of evading instances at training time, and show limited improvements in both test accuracy and robustness.
Gu et al.~\cite{gu2015} show preliminary results by augmenting deep neural networks with a pre-filtering layer 
based on a form of contractive auto-encoding. Most recently, Papernot et al.~\cite{papernot2015} shows the strong positive
effect of distillation on evasion robustness for neural networks. In this paper, we demonstrate a large increase in robustness 
for a boosted tree model hardened by \emph{adversarial boosting}. We empirically show that our method does not degrade accuracy and
creates the most robust model in our benchmark problem.

\section{The Optimal Evasion Problem}
In this section, we formally introduce the optimal evasion problem and briefly discuss its relevance to adversarial 
machine learning.
\label{sec:opteva}
We follow the definition of~\cite{biggio2013}. Let $c : \mathcal{X} \rightarrow \mathcal{Y}$ be a \underline{c}lassifier. 
For a given instance $x \in \mathcal{X}$ and 
a given ``distance'' function $d : \mathcal{X} \times \mathcal{X} \rightarrow \mathbb{R}_+$, the optimal evasion problem 
is defined as:
\begin{equation}
\label{problem:exact}
\underset{x' \in \mathcal{X}}{\text{minimize}} \; d(x, x') \quad \text{subject to} \; c(x) \neq c(x')
\end{equation}
 In this paper, we focus on binary classifiers defined over an $n$-dimensional feature space, that is
$\mathcal{Y}=\{-1, 1\}$ and $\mathcal{X} \subset \mathbb{R}^n$.

Setting the classifier $c$ aside, the distance function $d$ fully specifies~(\ref{problem:exact}), hence we 
talk about $d$-evading instances, or $d$-robustness. In fact, many problems of interest in adversarial machine learning 
fit under formulation~(\ref{problem:exact}) with a judicious choice for $d$. In the adversarial learning perspective, $d$ can
be used to model the cost the attacker has to pay for changing her initial instance $x$. In this paper, we proceed as if
this cost is decomposable over the feature dimensions. In particular, we present results for 
four representative distances. We briefly describe those and their typical effects on the solution of~(\ref{problem:exact}).

\paragraph{The $L_0$ distance} $\sum_{i=1}^n \mathbb{I}_{x_i \neq x'_i}$, or Hamming distance encourages the sparsest, 
most localized deformations with arbitrary magnitude. Our optimal evasion algorithm can also handle the case of 
non-uniform costs over features. This situation corresponds to minimizing $\sum_{i=1}^n \alpha_i \mathbb{I}_{x_i \neq x'_i}$
where $\alpha_i$ are non-negative weights.

\paragraph{The $L_1$ distance} $\sum_{i=1}^n |x_i - x'_i|$ encourages localized deformations 
and additionally controls for their magnitude.

\paragraph{The $L_2$ distance} $\sqrt{\sum_{i=1}^n (x_i - x'_i)^2}$ encourages less localized but 
small deformations.

\paragraph{The $L_\infty$ distance} $\max_i |x_i - x'_i|$ encourages uniformly spread deformations 
with the smallest possible magnitude.

Note that for binary-valued features, $L_1$ and $L_2$ reduce to $L_0$ and $L_\infty$ 
results in the trivial solution value 1 for~(\ref{problem:exact}).

\section{Evading Tree Ensemble Models}
We start by introducing tree ensemble models along with some useful notation. 
We then describe our optimal and approximate algorithms for generating evading instances on sum-ensembles of trees.


\subsection{Tree Ensembles}
\label{sub:tree_def}
A sum-ensemble of trees model $f:\mathbb{R}^n \rightarrow \mathbb{R}$ consists of a set $\mathcal{T}$ of regression trees. 
Without loss of generality, 
a regression tree $T \in \mathcal{T}$ is a binary tree where each internal node 
$n \in T.\text{nodes}$ holds a logical predicate $n.\text{predicate}$ over the feature variables, 
outgoing node edges are by convention labeled $n.\text{true}$ and $n.\text{false}$ and finally
each leaf $l \in T.\text{leaves}$ holds a numerical value $l.\text{prediction} \in \mathbb{R}$. For a given instance
$x \in \mathbb{R}^n$, the prediction path in $T$ is the path from the tree root to a leaf such that for each internal 
node $n$ in the path, $n.\text{true}$ is also in the path if and only if $n.\text{predicate}$ is true. The prediction of tree 
$T$ is the leaf value of the prediction path. Finally, the signed margin prediction $f(x)$ of the ensemble model is 
the sum of all individual tree predictions and the predicted label is obtained by thresholding, with the threshold value
commonly fixed at zero: $c(x)=1 \Leftrightarrow f(x) > 0$.

In this paper, we consider
the case of single-feature threshold predicates of the form $x_i < \tau$ or equivalently $x_i > \tau$, where 
$0 \leq i < n$ and $\tau \in \mathbb{R}$ are fixed model parameters. This restriction excludes oblique 
decision trees where predicates simultaneously involve several feature variables. 
We however note that oblique trees are 
seldom used in ensemble classifiers, partially because of their relatively high construction cost 
and complexity~\cite{norouzi2015}. 
Before describing our generic approach for solving the optimal evasion problem, we first state a simple 
worst-case complexity result for problem~(\ref{problem:exact}).

\subsection{Theoretical Hardness of Evasion}
\label{sub:reduction}
For a given tree ensemble model $f$, finding an $x \in \mathbb{R}^n$ such that $f(x)>0$ 
(or $f(x)<0$ without loss of generality) is NP-complete. That is, irrespectively of the choice for $d$, the optimal evasion
problem~(\ref{problem:exact}) requires solving an NP-complete feasibility subproblem. 

\begin{figure}[h]
        \centering
	\includegraphics[width=0.2\textwidth]{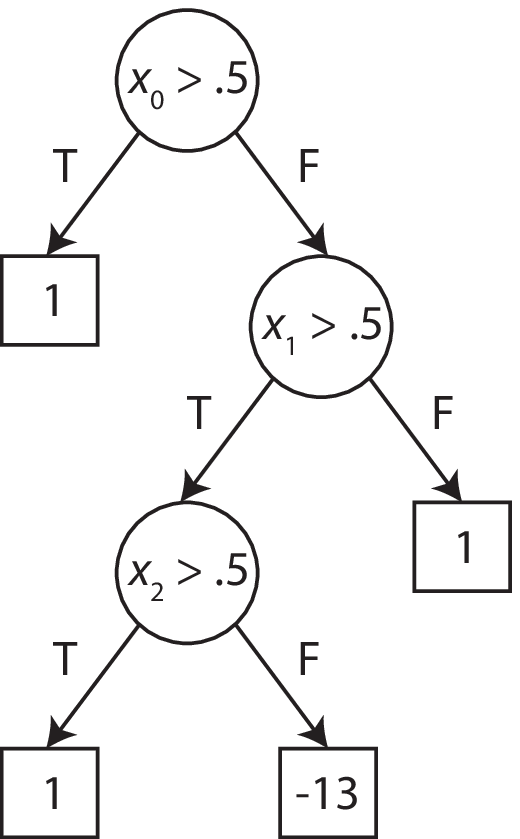}
	\caption{Regression tree for the clause $x_0 \vee \neg x_1 \vee x_2$. In this example,
$S$ has 13 clauses.}
	\label{fig:redux}
\end{figure}

We now give a proof of this fact by reduction from 3-SAT. 
First, given an instance $x$, computing the sign 
of $f(x)$ can be done in time at most proportional to the model size. 
Thus the feasibility problem is in NP. It is further NP-complete by a linear time reduction from 3-SAT as follows.
We encode in $x$ the assignment of values to the variables of the 3-SAT instance $S$. By convention, we choose $x_i > 0.5$ 
if and only if variable $i$ is set to true in $S$.
Next, we construct $f$ by arranging each clause of $S$ as a binary regression tree. 
Each regression tree has exactly one internal node per level, one for each variable appearing in the clause. 
Each internal node holds a predicate of the form $x_i > 0.5$ where $i$ is a clause variable. The nodes are arranged such
that there exists a unique prediction path corresponding to the falseness of the clause. For this path, the prediction 
value of the leaf is set to the opposite of the number of clauses in $S$, which is also the number of trees in the reduction.
The remaining leaves predictions are set to 1. Figure~\ref{fig:redux} illustrates this construction on an example.

It is easy to see that $S$ is satisfiable if and only if there exists $x$ such that $f(x)>0$. Indeed, 
a satisfying assignment for $S$ corresponds to $x$ such that $f(x)=|\mathcal{T}|>0$ and any 
non-satisfying assignment for $S$ corresponds to $x$ such that $f(x)\leq -1<0$ because there is at least one
false clause which corresponds to a regression tree which output is $-|\mathcal{T}|$.

While we can not expect an efficient algorithm for solving all instances 
of problem~(\ref{problem:exact}) unless $\text{P}=\text{NP}$, it may be the case that tree ensemble models as 
produced by common learners such as gradient boosting or random forests are practically easy to evade. 
We now turn to an algorithm for optimally solving the evasion problem when $d$ is one of the distances presented in 
section~\ref{sec:opteva}.

\subsection{Optimal Evasion}
\label{subsec:milp}
Let $f$ be a sum-ensemble of trees as defined in \ref{sub:tree_def} and $x \in \mathbb{R}^n$ an initial instance. 
We present a reduction of problem~(\ref{problem:exact}) into a Mixed Integer Linear Program (MILP). 
This reduction avoids introducing constraints with so called ``big-M'' constants~\cite{bigm}
at the cost of a slightly more complex solution encoding.
We experimentally find that our reduction produces tight formulations and acceptable running times for all common models $f$.

In what follows, we present the mixed integer program by defining three groups of MILP variables: 
the \textbf{predicate variables} encode the state (true or false) of all predicates, 
the \textbf{leaf variables} encode which prediction leaf is active in each tree, and
the optional \textbf{objective variable} for the case where $d$ is the $L_\infty$ norm.

We then introduce three families of constraints:
the \textbf{predicates consistency constraints} enforce the logical consistency between predicates,
the \textbf{leaves consistency constraints} enforce the logical consistency
between prediction leaves and predicates, and
the \textbf{model mislabel constraint} enforces the condition $c(x) \neq c(x')$, or 
equivalently that $f(x') > 0$ or $f(x') < 0$ depending on the sign of $f(x)$.
Finally we reduce the objective of~(\ref{problem:exact}) by relating the 
predicate variables to the value of $d(x, x')$ in \textbf{objective}.

\paragraph{Program Variables}
For clarity, MILP variables are \textbfit{bolded and italicized} throughout. Our reduction uses three families of variables.
\begin{itemize}
\item At most $\sum_{T \in \mathcal{T}} |T.\text{nodes}|$ binary variables $\textbfit{p}_i \in \{0;1\}$ 
(\underline{p}redicates) encoding the state of the predicates. 
Our implementation sparingly create those variables: if
any two or more predicates in the model are logically equivalent, 
their state is represented by a single variable. 
For example, the state of $x'_5 < 0$ and $-x'_5 > 0$ would be represented by the same variable.

\item $\sum_{T \in \mathcal{T}} |T.\text{leaves}|$ continuous variables $0 \leq \textbfit{l}_i \leq 1$ 
(\underline{l}eaves) encoding which prediction leaf is active in each tree. The MILP constraints 
force exactly one $\textbfit{l}_i$ per tree to be non-zero with $\textbfit{l}_i = 1$. The 
$\textbfit{l}$ variables are thus implied binary in any solution but are nonetheless typed 
continuous to narrow down
the choice of branching variable candidates during branch-and-bound, 
and hence improve solving time.

\item At most 1 non-negative continuous variable $\textbfit{b}$ (\underline{b}ound) for expressing the 
distance $d(x, x')$ of problem~(\ref{problem:exact}) when $d$ is the $L_\infty$ distance. 
This variable is first used in the \emph{objective} paragraph.
\end{itemize}

\begin{figure}[t]
        \centering
	\includegraphics[width=0.23\textwidth]{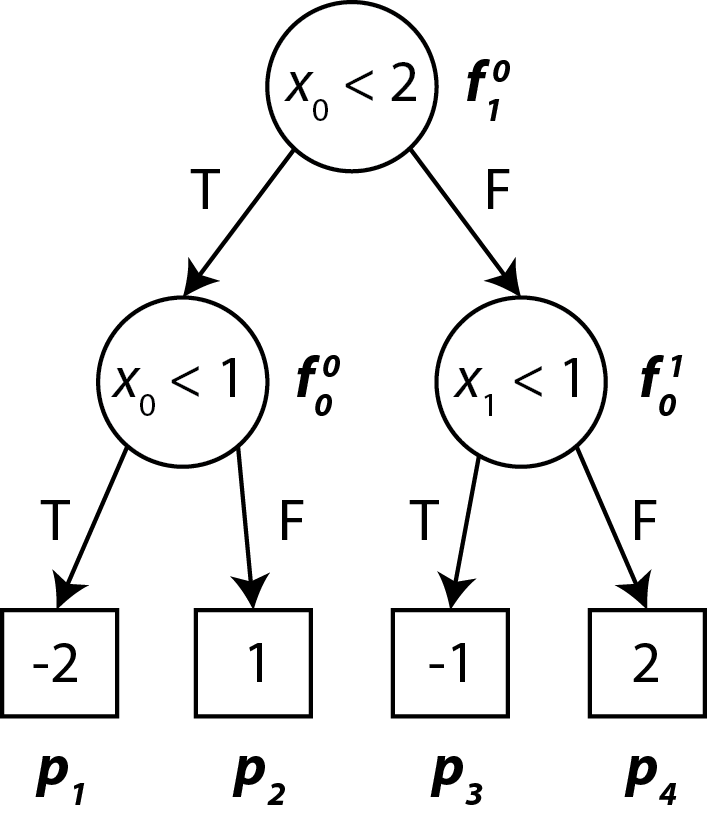}
	\caption{Regression tree for the reduction example. Predicate variables $\textbfit{p}$ and 
leaf variables $\textbfit{l}$ are shown next to their corresponding internal and leaf nodes. There are 
 $n=2$ continuous features. The leaf predictions are -2, 1, 1 and 2.}
	\label{fig:tree_example}
\end{figure}

In what follows, we illustrate our reduction by using a model with a single 
regression tree as represented in figure~\ref{fig:tree_example}. 

\paragraph{Predicates consistency}
Without loss of generality, each predicate variable $\textbfit{p}_i$ corresponds to the state of a predicate 
of the form $x_k < \tau_k$. If two variables $\textbfit{p}_i$ and $\textbfit{p}_j$ correspond to predicates over the same
variable $x_k < \tau_1$ and $x_k < \tau_2$, then $\textbfit{p}_i$ and $\textbfit{p}_j$ can take inconsistent 
values without additional constraints. For instance, if $\tau_1 < \tau_2$, 
then $\textbfit{p}_i = 1$ and $\textbfit{p}_j = 0$ would be logically inconsistent because 
$x_k < \tau_1 \Rightarrow x_k < \tau_2$, but any other valuation is possible.

For each feature variable $x'_k$, we can ensure the consistency of all $\textbfit{p}$ variables which reference 
a predicate over $x'_k$ by adding $K-1$ inequalities enforcing the implicit implication constraints between 
the predicates, where $K$ is the number of $\textbfit{p}$ variables referencing $x_k$.
For a given $x'_k$, let $\tau_1 < \dots < \tau_K$ be the sorted thresholds of the predicates over $x'_k$. 
Let $\textbfit{p}_1, \dots, \textbfit{p}_K$ be the MILP variables corresponding to predicates $x'_k < \tau_1, \dots, x'_k < \tau_K$. A valuation of $(\textbfit{p}_i)_{i=1..K}$ is consistent if and only 
if $\textbfit{p}_1 = 1 \Rightarrow \dots \Rightarrow \textbfit{p}_K = 1$. Thus the consistency constraints are:
\[
\textbfit{p}_1 \leq \textbfit{p}_2 \leq \dots \leq \textbfit{p}_K
\]

When the feature variables $x'_k$ are binary-valued, there is a single $\textbfit{p}_i$ variable associated to a feature 
variable: all predicates $x'_k < \tau$ with $0 < \tau < 1$ are equivalent. Generally, tree building packages generate
a threshold of $0.5$ in this situation. This is however implementation dependent and we 
can simplify the formulation with additional 
knowledge of the value domain $x'_k$ is allowed to take.

In our toy example in figure~\ref{fig:tree_example}, variables $\textbfit{p}_0$ and $\textbfit{p}_1$ refer to the same feature dimension
$0$ and are not independent. The predicate consistency constraint in this case is:
\[
\textbfit{p}_1 \leq \textbfit{p}_0
\]
and no other predicate consistency constraint is needed.

\paragraph{Leaves consistency} These constraints bind the $\textbfit{p}$ and $\textbfit{l}$ variables so that the semantics 
of the regression trees are preserved. Each regression tree has its own independent set of leaves consistency constraints. 
We construct the constraints such that the following properties hold:
\begin{enumerate}[label=(\roman*)]
\item if $\textbfit{l}_k = 1$, then every other $\textbfit{l}_{i \neq k}$ variable within the same tree is zero, and
\item if a leaf variable $\textbfit{l}_k$ is 1, then all predicate variables $\textbfit{p}_i$ encountered in the prediction path
of the corresponding leaf are forced to be either 0 or 1 in accordance with the semantics of the prediction path, and
\item exactly one $\textbfit{l}_k$ variable per tree is equal to 1. This property is needed 
because (i) does not force any $\textbfit{l}_i$ to be non-zero.
\end{enumerate}

Enforcing property (i) is done using a classic exclusion constraint. 
If $\textbfit{l}_1, \dots, \textbfit{l}_K$ are the $K$ leaf variables for
a given tree, then the following equality constraint enforces (i):
\begin{equation}
\label{constr:exclusion}
\textbfit{l}_1 + \textbfit{l}_2 + \dots + \textbfit{l}_K = 1
\end{equation}

For our toy example, this constraint is:
\[
\textbfit{l}_1 + \textbfit{l}_2 + \textbfit{l}_3 + \textbfit{l}_4 = 1
\]

Enforcing property (ii) requires two constraints per internal node. Let us start at the root node $r$. 
Let $\textbfit{p}_{\text{root}}$ 
be the variable corresponding to the root predicate. Let $\textbfit{l}^T_1, \dots, \textbfit{l}^T_i$ be the 
variables corresponding to the leaves of the subtree rooted at $r.\text{true}$, and 
$\textbfit{l}^F_1, \dots, \textbfit{l}^F_j$ the variables for the subtree rooted at $r.\text{false}$. The root predicate is true if
and only if the active prediction leaf belongs to the subtree rooted at $r.\text{true}$. In terms of the MILP reduction, 
this means that $\textbfit{p}_{\text{root}}$ is equal 
to 1 if and only if one of the leaf variables of the true subtree is set to one. Similarly on the false subtree, $\textbfit{p}_{\text{root}}$ is 0 if and only if one of the leaf variables of the false subtree is set to one. Because only one leaf can be non-zero, 
these constraints can be written as:
\[
 1 - \left(\textbfit{l}^F_1 + \textbfit{l}^F_2 + \dots + \textbfit{l}^F_j
 \right) = \textbfit{p}_{\text{root}} = \textbfit{l}^T_1 + \textbfit{l}^T_2 + \dots + \textbfit{l}^T_i
\]

The case of internal nodes is identical, except that if and only ifs are weakened to single side implications. 
Indeed, unlike the root case, it is possible that no leaf in either subtree might be an active prediction leaf.
For an internal node $n$, let $\textbfit{p}_{\text{node}}$ be the variable attached to the node, 
$\textbfit{l}^T$ and $\textbfit{l}^F$ the variables attached to leaves of the true and false subtrees rooted at 
$n.\text{true}$ and $n.\text{false}$. The constraints are:
\[
1 - \left( \textbfit{l}^F_1 + \textbfit{l}^F_2 + \dots + \textbfit{l}^F_j \right) \geq \textbfit{p}_{\text{node}} \geq \textbfit{l}^T_1 + \textbfit{l}^T_2 + \dots + \textbfit{l}^T_i
\]

In our toy example, we have 3 internal nodes and thus six constraints. The constraints associated with the root, the leftmost 
and rightmost internal nodes are respectively:
\begin{align*}
\textbfit{l}_1 + \textbfit{l}_2 &= \textbfit{p}_0 = 1 - \left( \textbfit{l}_3 + \textbfit{l}_4 \right)\\
\textbfit{l}_1 &\leq \textbfit{p}_1 \leq 1 - \textbfit{l}_2 \\
\textbfit{l}_3 &\leq \textbfit{p}_2 \leq 1 - \textbfit{l}_4
\end{align*}

Finally, property (iii) automatically holds given the previously defined constraints. To see this, one can walk down the 
prediction path defined by the $\textbfit{p}$ variables and notice that at each level, the leaves values of one of 
the subtree rooted at the current node must be all zero. For instance, if $\textbfit{p}_{\text{node}} = 1$, then we have
\begin{align*}
\textbfit{l}^F_1 + \textbfit{l}^F_2 + \dots + \textbfit{l}^F_j \leq 0
\Rightarrow \textbfit{l}^F_1 = \textbfit{l}^F_2 = \dots = \textbfit{l}^F_j = 0
\end{align*}
At the last internal node, exactly two leaf variables remain unconstrained, and one of them is pushed to zero. 
By the exclusion constraint~(\ref{constr:exclusion}), the remaining leaf variable must be set to 1.

\paragraph{Model mislabel} Without loss of generality, consider an original instance $x$ such that $f(x) < 0$. In order for
$x'$ to be an evading instance, we must have $f(x') \geq 0$. Encoding the model output $f(x')$ is straightforward given the
leaf variables $\textbfit{l}$. The output of each regression tree is simply the weighted sum of its leaf variables, where the
weight of each variable $\textbfit{l}_i$ corresponds to the prediction value $v_i$ of the associated leaf. 
Hence, $f(x')$ is the sum of $|\mathcal{T}|$ weighted sums over the $\mathbf{l}$ variables and 
the following constraint enforces
$f(x') \geq 0$:
\[
\sum_{i} v_i \textbfit{l}_i \geq 0
\]

For our running example, the mislabeling constraint is:
\[
-2\textbfit{l}_1 + \textbfit{l}_2 - \textbfit{l}_3 + 2\textbfit{l}_4 \geq 0
\]

\paragraph{Objective}
Finally, we need to translate the objective $d(x, x')$ of problem~(\ref{problem:exact}). 
We rely on the predicate variables $\textbfit{p}$ in doing so. For any distance $L_\rho$ with $\rho \in \mathbf{N}$,
there exists weights $w_i$ and a constant $C$ such that the MILP objective can be written as:
\[
\sum_{i} w_i \textbfit{p}_i + C
\]

We now describe the construction of $(w_i)_i$ and $C$. Recall that for each feature dimension $1 \leq k \leq n$, 
we have a collection of predicate variables $(\textbfit{p}_i)_{i=1..K}$ associated with predicates 
$x'_k < \tau_1, \dots, x'_k < \tau_K$ where the thresholds are sorted $\tau_1 < \dots < \tau_K$. 
Thus, the $\textbfit{p}$ variables effectively encode the interval to which $x'_k$ belongs to, and any feature 
value within the interval will lead to the same prediction $f(x')$. 
There are exactly $K+1$ distinct possible valuations for the binary variables 
$\textbfit{p}_1 \leq \textbfit{p}_2 \leq \dots \leq \textbfit{p}_K$ and the value domain mapping 
$\phi : \textbfit{p} \rightarrow \left(\mathbf{R} \cup \{-\infty; \infty\}\right)^2$ is:
\begin{align*}
x'_k &\in \phi(\textbfit{p}) = [\tau_i, \tau_{i+1}) \\
     &i = \max \{k | \textbfit{p}_k = 0, 0 \leq k \leq K+1\}
\end{align*}
where by convention $\textbfit{p}_0 = 0$, $\textbfit{p}_{K+1} = 1$ and $\tau_0 = -\infty$, $\tau_{K+1} = \infty$.
Setting aside the $L_\infty$ case for now, consider $\rho \in \mathbb{N}$ the norm we are interested in for $d$. 
Instead of directly minimizing $\|x-x'\|_\rho$, our formulation equivalently minimizes $\|x-x'\|^\rho_\rho$. 
By minimizing the latter, we are able to consider the contributions of each feature dimension independently:
\[
\|x-x'\|^\rho_\rho = \sum_{k=1}^n |x_k - x'_k|^\rho
\]
We take $0^0 = 0$ by convention.
At the optimal solution, $|x_k - x'_k|^\rho$ can only take $K+1$ distinct values. Indeed, if $x'_k$ and $x_k$ belong
to the same interval, then $x'_k = x_k$ minimizes the distance along feature $k$, and this distance is
zero. If $x'_k$ and $x_k$ do not belong to same interval, then setting $x'_k$ at the border of $\phi(\textbfit{p})$
that is closest to $x_k$ minimizes the distance along $k$. 
If $\phi(\textbfit{p}) = [\tau_i, \tau_{i+1})$, this distance is simply
equal to $\min \{|x_k-\tau_i|^\rho, |x_k - \tau_{i+1}|^\rho\}$. Note that because of the right-open interval, 
the minimum distance is actually an infimum. In our implementation, we simply use a guard value $\epsilon = 10^{-4}$ 
of the same magnitude order than the numerical tolerance of the MILP solver.

Hence, we can express the minimization objective of problem~(1) as a weighted sum of $\textbfit{p}$
 variables without loss of generality. 
Let $0 \leq j \leq K+1$ be the indices such that $x_k \in [\tau_j, \tau_{j+1})$. Let $(w_i)_{i=0..K+1}$ 
such that for any valid valuation of $\textbfit{p}$ we have
$\sum_{i=0}^{K+1} w_i \textbfit{p}_i = \inf_{x'_k \in \phi(\textbfit{p})} |x_k - x'_k|^\rho$. By the discussion above and 
exhaustively enumerating the $K+1$ valuations of $\textbfit{p}$, $w$ is the
solution to the following $K+1$ equations:
\begin{align*}
\label{eqn:system}
w_{K+1} &= | x_k - \tau_K|^\rho \\
w_{K} + w_{K+1} &= | x_k - \tau_{K-1}|^\rho \\
&\dots \\
w_{j+1} + \dots + w_{K+1} &= |x_k - \tau_{j+1}|^\rho \\
w_j + w_{j+1} + \dots + w_{K+1} &= 0 \\
w_{j-1} + w_{j} + w_{j+1} + \dots + w_{K+1} &= |x_k - \tau_{j} - \epsilon|^\rho \\
&\dots \\
w_1 + w_2 + w_3 + \dots + w_{K+1} &= |x_k - \tau_2 - \epsilon|^\rho\\
w_0 + w_1 + w_2 + w_3 + \dots + w_{K+1} &= |x_k - \tau_1 - \epsilon|^\rho
\end{align*}
Note that this system of linear equations is already in triangular form and obtaining the $w$ values is immediate. 
To obtain the full MILP objective, we repeat this process for every feature $1 \leq k \leq n$ and take the sum of
all weighted sums of subsets of $\textbfit{p}$.

Finally, for the $L_\infty$ case, we use 1 continuous variable $\textbfit{b}$. We introduce $n$ additional constraints
to the formulation, one for each feature dimension $k$. As per the previous discussion, we can generate the weights $w$ such
that $\sum_{i=0}^{K+1} w_i \textbfit{p}_i = \inf_{x'_k \in \phi(\textbfit{p})} |x_k - x'_k|$ (this is the $\rho=1$ case). The
additional constraint on dimension $k$ is then:
\[
\sum_{i=0}^{K+1} w_i \textbfit{p}_i \leq \textbfit{b}
\]
and the MILP objective is simply the variable $\textbfit{b}$ itself.

For our toy example, consider $(x_0=0, x_1=3)$. In the case of the $L_0$ distance, we have the following objective:
\[
1 - \textbfit{p}_1 + \textbfit{p}_2
\]
For the (squared) $L_2$ distance instead, the objective is essentially:
\begin{align*}
4 - 3\textbfit{p}_0 -\textbfit{p}_1 + 4\textbfit{p}_2
\end{align*}
For the $L_\infty$ case, our objective reduces to the variable $\textbfit{b}$ and we introduce $n$ additional 
bounding constraints of the form $\dots \leq \textbfit{b}$ where the left hand side measures $|x_k - x'_k|$ using the
same technique as the $\rho=1$ case.

Hence, the full MILP reduction of the optimal $L_0$-evasion for our toy instance is:
\begin{align*}
\min_{\textbfit{p},\textbfit{l}} \; &1 - \textbfit{p}_1 + \textbfit{p}_2 \\
\text{s.t.} \;\; &\textbfit{p}_0, \textbfit{p}_1 \in \{0;1\}; 0 \leq \textbfit{l}_1, \textbfit{l}_2, \textbfit{l}_3, \textbfit{l}_4 \leq 1 \\ 
&\textbfit{p}_1 \leq \textbfit{p}_0 \tag*{predicates consistency}\\
&\textbfit{l}_1 + \textbfit{l}_2 + \textbfit{l}_3 + \textbfit{l}_4 = 1 \tag*{leaves consistency}\\
&\textbfit{l}_1 + \textbfit{l}_2 = \textbfit{p}_0 = 1 - \left( \textbfit{l}_3 + \textbfit{l}_4 \right)\tag*{leaves consistency}\\
&\textbfit{l}_1 \leq \textbfit{p}_1 \leq 1 - \textbfit{l}_2 \tag*{leaves consistency}\\
&\textbfit{l}_3 \leq \textbfit{p}_2 \leq 1 - \textbfit{l}_4 \tag*{leaves consistency}\\
&-2\textbfit{l}_1 + \textbfit{l}_2 - \textbfit{l}_3 + 2\textbfit{l}_4 \geq 0 \tag*{model mislabel}
\end{align*}

\paragraph{Additional Constraints}
Reducing problem~(\ref{problem:exact}) to a MILP allows expressing potentially complex inter-feature
dependencies created by the feature extraction step. For instance, consider the common case of 
$K$
mutually exclusive binary features $x_1, \dots, x_K$ such that in any well-formed instance, 
exactly one feature is non-zero.
Letting $\textbfit{p}_i$ be the predicate variable associated with $x_i < 0.5$, mutual exclusivity
can be enforced by:
\[
\sum_{i=1}^{K} \textbfit{p}_i = K - 1
\]

\subsection{Approximate Evasion}
\label{sub:greedy}
While the above reduction of problem~(\ref{problem:exact}) to an MILP is linear in the size of the model $f$, the 
actual solving time can be very significant for difficult models. Thus, as a complement to the exact method, we 
develop an approximate evasion algorithm to generate good quality evading instances. For this part, we exclusively 
focus on minimizing the $L_0$ distance. Our approximate evasion algorithm is based on the iterative 
coordinate descent procedure described in algorithm~\ref{alg:coordinate_descent}. 
\begin{algorithm}[h]
   \caption{Coordinate Descent for Problem~(\ref{problem:exact})}
   \label{alg:coordinate_descent}
\begin{algorithmic}
   \STATE {\bfseries Input:} model $f$, initial instance $x$ (assume $f(x) < 0$)
   \STATE {\bfseries Output:} evading instance $x'$ such that $f(x')\geq0$
   \STATE \vspace{-1em} 
   \STATE $x' \gets x$
   \WHILE {$f(x') < 0$}\vspace{-1em}
   \STATE {\[x' \gets \argmax_{\tilde x': \|\tilde x' - x'\|_0 = 1} f(\tilde x')\]}\vspace{-1em}
   \ENDWHILE
\end{algorithmic}
\end{algorithm}

In essence, this algorithm greedily modifies the single best feature at each iteration until the sign of $f(x')$ changes. 
We now present an efficient algorithm for solving the inner optimization subproblem
\begin{equation}
\label{problem:discrete}
\max_{\tilde x: \|x - \tilde x\|_0 = 1} f(\tilde x)
\end{equation}

The time complexity of a careful brute force approach is high. 
For balanced regression trees, the prediction time for a given 
instance is $O\left(\sum_{T \in \mathcal{T}} \log |T.\text{nodes}|\right)$. 
Further, for each dimension $1 \leq k \leq n$, we must compute all
possible values of $f(\tilde x)$ where $\tilde x$ and $x$ only differ along dimension $k$. 
Note that because the model predicates effectively discretize the feature space, 
$f(\tilde x)$ takes a finite number of distinct values. This number is no 
more than one plus the total number of predicates holding over feature $k$. 
Hence, we must compute $f(\tilde x)$ for a total of $\sum_{T \in \mathcal{T}} |T.\text{nodes}|$ candidates $\tilde x$ 
and the total running time is $O\left(\sum_{T \in \mathcal{T}} |T.\text{nodes}| \times \sum_{T \in \mathcal{T}} \log |T.\text{nodes}|\right)$. If we denote by $|f|$ the size of the model which is proportional to the total number of predicates, 
the running time is $O\left(|f||\mathcal{T}|\log \frac{|f|}{|\mathcal{T}|}\right)$. Tree ensembles often have thousands of
trees, making the $|f||\mathcal{T}|$ dependency prohibitively expensive.

We can efficiently solve problem~(\ref{problem:discrete}) by a dynamic programming approach. The main idea is to
visit each internal node no more than once by computing what value of $\tilde x$ can land us at each node. 
We call this approach \emph{symbolic prediction} in reference to 
symbolic program execution~\cite{king1976}, because we essentially move a symbolic
instance $\tilde x$ down the regression tree and keep track of the constraints imposed on $\tilde x$ 
by all encountered predicates. Because 
we are only interested in $\tilde x$ instances that are at most one feature away from $x$, we can stop the tree exploration
early if the current constraints imply that at least two dimensions need to be modified or more trivially, if there is no 
instance $\tilde x$ that can simultaneously satisfy all the constraints. When reaching a leaf, we report
 the leaf prediction value $f(\tilde x)$ along with the pair of perturbed dimension number $k$ and 
value interval for $\tilde x_k$ which would reach the given leaf.

To simplify the presentation of the algorithm, we introduce a \textsc{SymbolicInstance} data structure which keeps track of
the constraints on $\tilde x$. This structure is initialized by $x$ and has four methods. 
\begin{itemize}
\item For a predicate $p$, .\textsc{isFeasible}($p$)  returns true if and only if there exists an instance $\tilde x$ 
such that $\|\tilde x - x\|_0 \leq 1$ and all constraints including $p$ hold.
\item .\textsc{update}($p$)  updates the set of constraints on $\tilde x$ by adding predicate $p$.
\item .\textsc{isChanged}()  returns true if and only if the current set of constraints imply $x \neq \tilde x$.
\item .\textsc{getPerturbation}()  returns the index $k$ such that $x_k \neq \tilde x_k$ and the admissible interval 
of values for $\tilde x_k$
\end{itemize}
It is possible to implement \textsc{SymbolicInstance}  such that each method executes in constant time.

Algorithm~\ref{alg:sympred} presents the symbolic prediction algorithm recursively for a given tree. 
It updates a list of elements by appending tuples to it. 
The first element of a tuple is the feature index $k$ where $\tilde x_k \neq x_k$, the second element is the allowed
right-open interval for $\tilde x_k$, and the last element is the prediction score $f(\tilde x)$.

\begin{algorithm}[h]
   \caption{Recursive definition of the symbolic prediction algorithm. For the first call,
   $n$ is the tree root, $s$ is a fresh \textsc{SymbolicInstance} object initialized 
  on $x$ with no additional constraints and $l$ is an empty list.}
   \label{alg:sympred}
\begin{algorithmic}
   \STATE {\bfseries Input:} node $n$ (either internal or leaf)
   \STATE {\bfseries Input:} $s$ of type \textsc{SymbolicInstance}
   \STATE {\bfseries Input/Output:} list of tuples $l$ (see description)
   \STATE \vspace{-1em}
   \IF {$n$ is a leaf}
      \IF {$s.\textsc{isChanged}()$}
         \STATE $l \gets l \cup $\{$s.\textsc{getPerturbation}()$, $n.\text{prediction}$\}
      \ENDIF
   \ELSE
      \IF {$s.\textsc{isFeasible}(n.\text{predicate})$}
	\STATE $s_T \gets \textsc{copy}(s)$
	\STATE $s_T.\textsc{update}(n.\text{predicate})$
        \STATE \textsc{symbolicPrediction}($n.\text{true}$, $s_T$, $l$)
      \ENDIF
      \IF {$s.\textsc{isFeasible}(\neg n.\text{predicate})$}
	\STATE $s.\textsc{update}(\neg n.\text{predicate})$
        \STATE \textsc{symbolicPrediction}($n.\text{false}$, $s$, $l$)
      \ENDIF
   \ENDIF
   \end{algorithmic}
\end{algorithm}
This algorithm visits each node at most once and performs at most one copy of the \textsc{SymbolicInstance} $s$ per visit. 
The copy operation is proportional to the number of constraints in $s$. For a balanced tree $T$, the copy cost is 
$O(\log |T.\text{nodes}|)$, so that the total running time is $O(|T.\text{nodes}| \log |T.\text{nodes}|)$.

For each tree of the model, once the list of dimension-interval-prediction tuples is obtained, 
we substract the leaf prediction value for $x$ from all predictions in order to obtain a 
score variation between $\tilde x$ and $x$ instead of the score for $\tilde x$. With the help of an additional 
data structure, we can use the dimension-interval-variation tuples across all trees
to find the dimension $k$ and interval for $\tilde x_k$ which corresponds to the highest variation 
$f(\tilde x) - f(x)$. This final search can be done in $O(L \log L)$, where $L$ is the total number of tuples,
and is no larger than $\sum_{T \in \mathcal{T}} |T.\text{leaves}|$ by construction. To summarize, the time complexity of 
our method for solving problem~(\ref{problem:discrete}) is $O(|f| \log |f|)$, an exponential improvement over the brute force
method.

\section{Results}
We turn to the experimental evaluation of the robustness of tree ensembles. 
We start by describing the evaluation dataset and our choice of models for benchmarking 
purposes before moving to a 
quantitative comparison of the robustness of boosted trees and random forest models against a 
garden variety of learning algorithms. We finally show that the brittleness of boosted trees can be 
effectively addressed by including fresh evading instances in the training set during boosting.

\begin{figure*}[t]
        \centering
        \begin{subfigure}[b]{0.245\textwidth}
                \includegraphics[width=\textwidth]{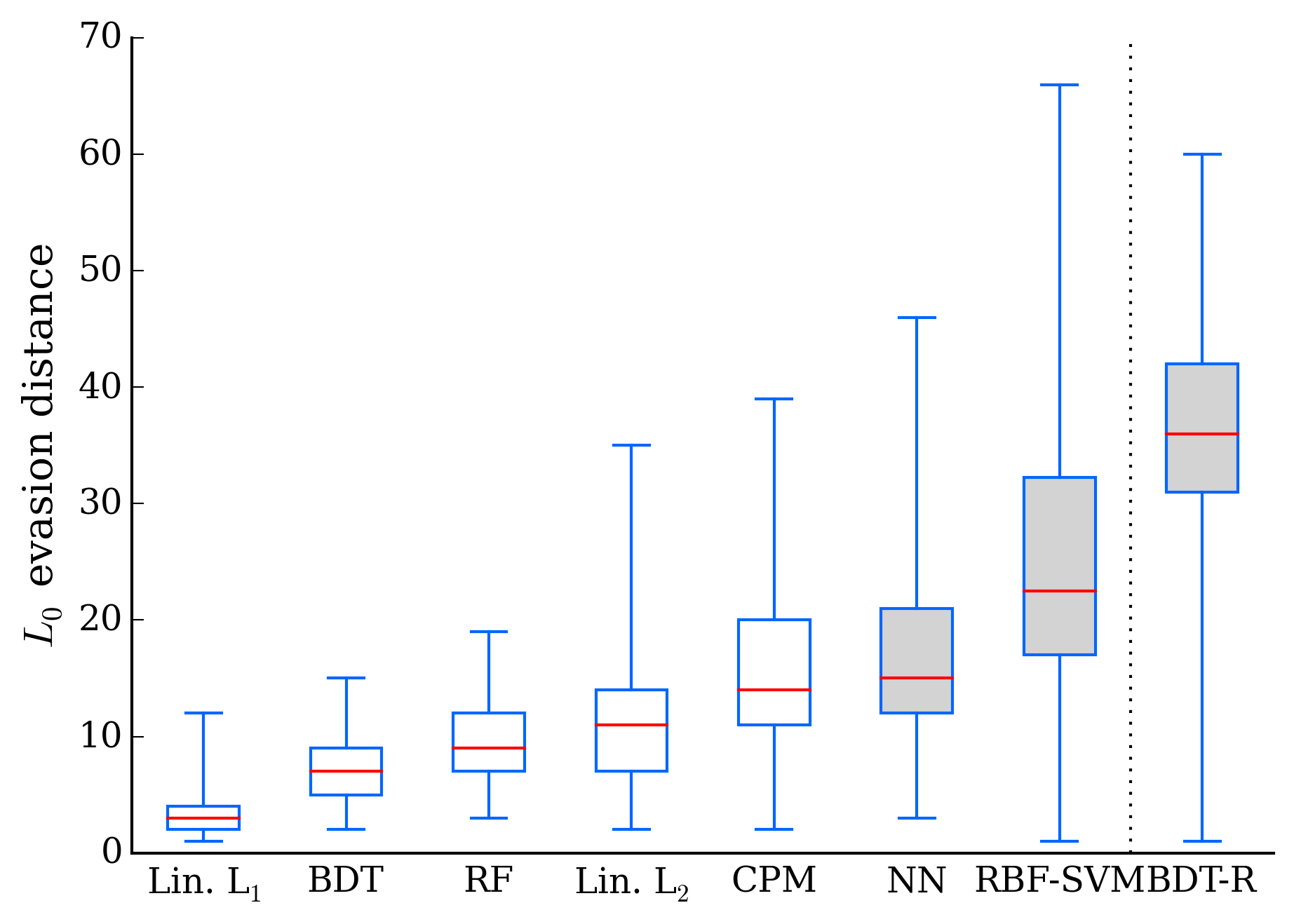}
        \end{subfigure}
        \begin{subfigure}[b]{0.245\textwidth}
                \includegraphics[width=\textwidth]{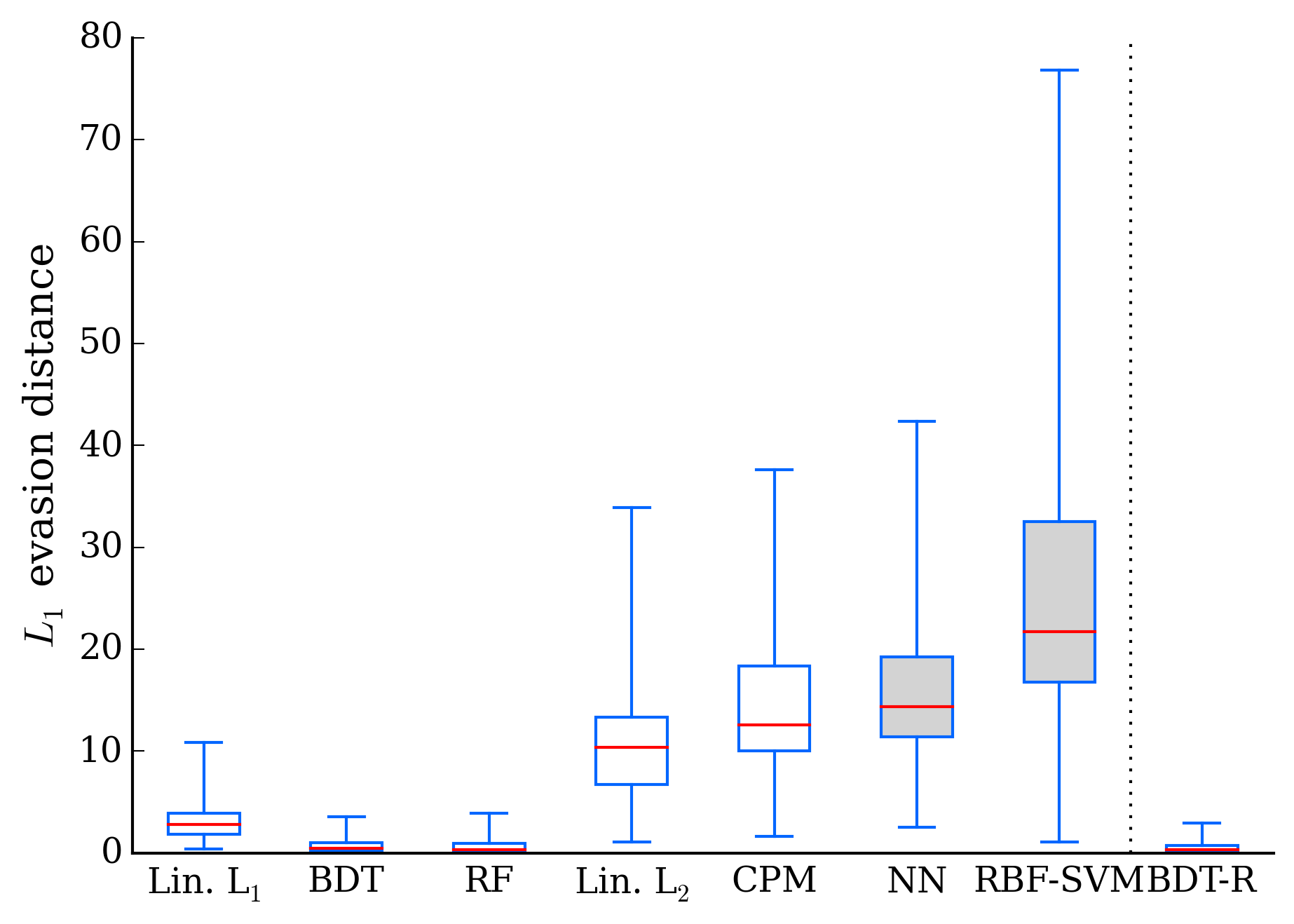}
        \end{subfigure}
        \begin{subfigure}[b]{0.245\textwidth}
                \includegraphics[width=\textwidth]{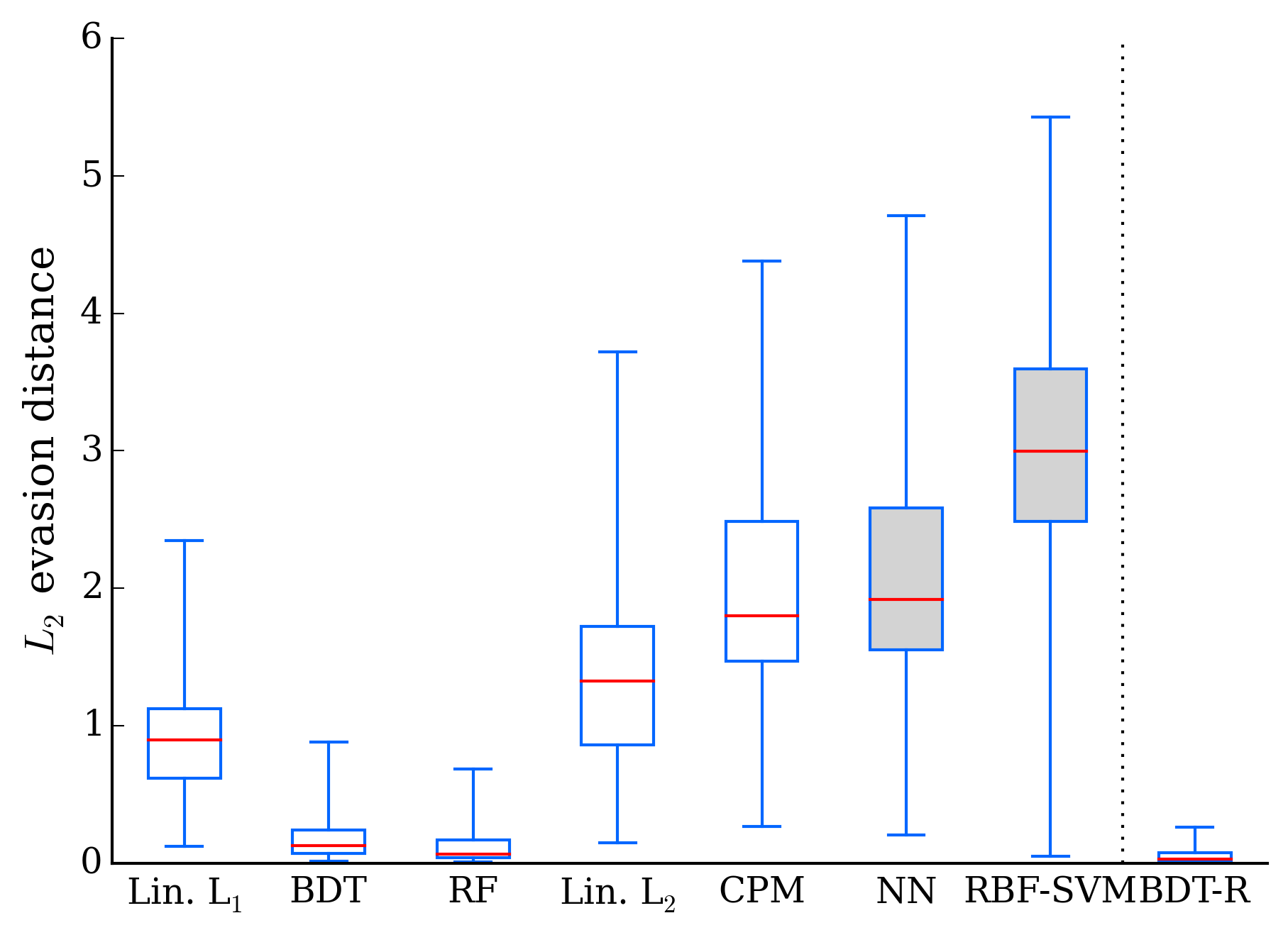}
        \end{subfigure}
        \begin{subfigure}[b]{0.245\textwidth}
		 \includegraphics[width=\textwidth]{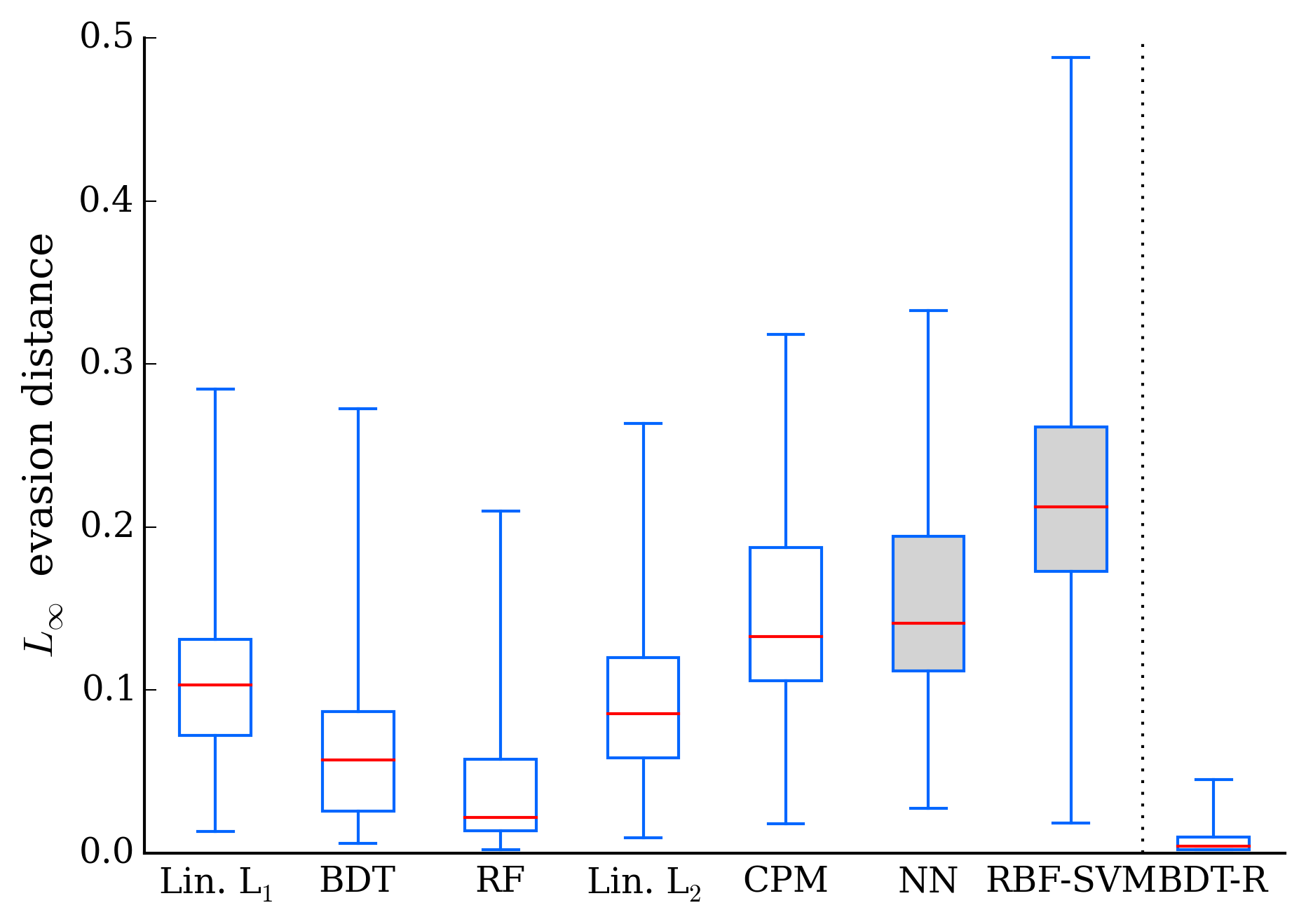}
        \end{subfigure}
        \caption{Optimal (white boxes) or best-effort (gray boxes) evasion bounds for different metrics 
on the evaluation dataset. 
		The smallest bounds, 25-50\% and 50-75\% quartiles and largest bounds are shown. 
                The red line is the median score. Larger scores mean more deformations are necessary to
		change the model prediction.}\label{fig:comp}
\vspace{-1.5em}
\end{figure*}

\begin{table}[h]
\begin{tabular}{l | l | l}
Model & Parameters & Test Error \\
\hline
\hline
Lin. $L_1$ & $C=0.5$ & 1.5\% \\
Lin. $L_2$ & $C=0.2$ & 1.5\% \\
BDT & 1,000 trees, depth 4, $\eta=0.02$ & 0.25\% \\
RF & 80 trees, max. depth 22 & 0.20\% \\
CPM & $k=30$, $C=0.01$& 0.20\% \\
NN & 60-60-30 sigmoidal (tanh) units& 0.25\% \\
RBF-SVM & $\gamma=0.04$, $C=1$& 0.25\% \\
\hline
BDT-R & 1,000 trees, depth 6, $\eta=0.01$& 0.20\%
\end{tabular}
\caption{\label{table:models}The considered models. BDT-R is the hardened boosted trees model introduced in section~\ref{subsec:hardening}.}
\end{table}


\subsection{Dataset and Method}
We choose digit recognition over the MNIST~\cite{mnist} dataset as our benchmark 
classification task for three reasons. 
First, the MNIST dataset is well studied and exempt from labeling errors. 
Second, there is a one-to-one mapping between pixels and features, so that features can vary independently from each other.
Third, we can pictorially represent evading instances, and this helps understanding the models' robustness or lack of.
Our running binary classification task is to distinguish between handwritten digits ``2'' and ``6''. 
Our training and testing sets respectively include 11,876 and 1,990 images and each image has $28\times28$ gray scale pixels and 
our feature space is $\mathcal{X} = [0,1]^{784}$. As our main goal is not to compare model accuracies,
but rather to obtain the best possible model for each model class, we tune the hyper-parameters so as 
to minimize the error on the testing set directly.
In addition to the training and testing sets, we
create an evaluation dataset of a hundred instances from the testing set such that every 
instance is correctly
classified by \emph{all} of the benchmarked models. These correctly classified instances are to serve the purpose of
$x$, the starting point instances in the evasion problem~(\ref{problem:exact}).

\subsection{Considered Models}
Table~\ref{table:models} summarizes the 7 benchmarked models with their salient hyper-parameters
and error rates on the testing set. For our 
tree ensembles, BDT is a (gradient) boosted decision trees model in the modern XGBoost 
implementation~\cite{xgboost} and RF is a random forest trained using scikit-learn~\cite{sklearn}. 
We also include the following models for comparison purposes. Lin. $L_1$ and Lin. $L_2$ are 
respectively a $L_1$ and 
$L_2$-regularized logistic regression using the LibLinear~\cite{liblinear} implementation. 
RBF-SVM is a regular Gaussian kernel SVM trained using LibSVM~\cite{libsvm}. NN is a 3 hidden layer
neural network with a top logistic regression layer implemented using Theano~\cite{theano} (no pre-training, no drop-out).
Finally, our last benchmark model is the equivalent of a shallow 
neural network made of two max-out units (one unit for each class) each made of thirty linear classifiers. This model 
corresponds to the difference of two Convex Polytope Machines~\cite{cpm} (one for each class) and we use the authors' 
implementation (CPM). Two factors motivate the choice of CPM. 
First, previous work has theoretically considered the evasion robustness of such ensemble of linear 
classifiers and proved the problem to be NP-hard~\cite{stevens2013}. Second, unlike RBF-SVM and NN, this 
model can be readily reduced to a Mixed Integer Program, enabling optimal evasions thanks to a MIP solver. As the reduction 
is considerably simpler than the one presented for tree ensembles above, we omit it here. 
Except for the two linear classifiers, all models have a comparable, very low error rate on 
the benchmark task.

\begin{figure}[h]
\centering
  \includegraphics[width=0.47\textwidth]{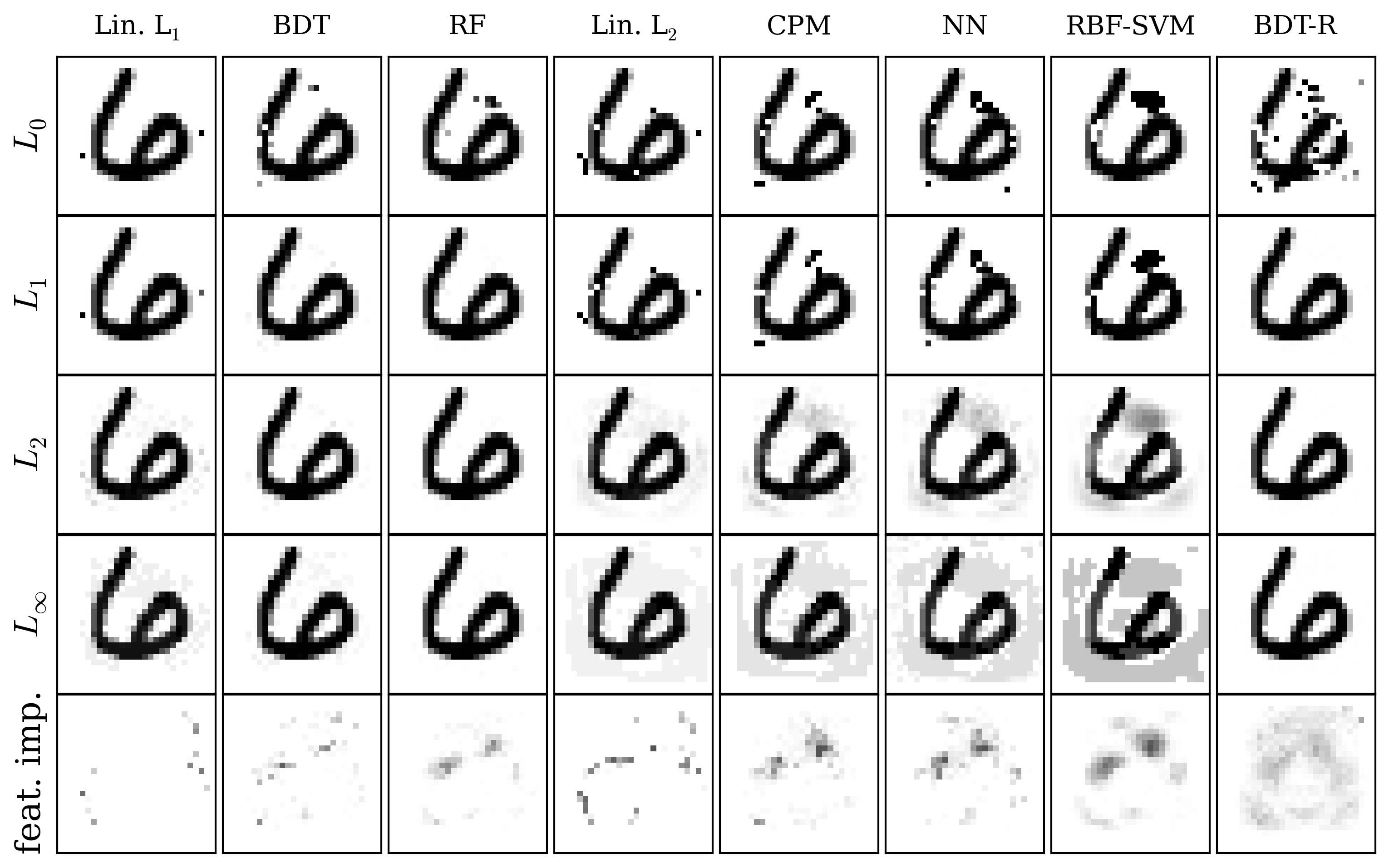}
  \caption{\label{fig:perfs_q} First 4 rows: examples of optimal or best effort evading ``6'' instances. 
Every picture is misclassified as ``2'' by its column model. Last row: feature importance computed as frequency
of pixel modification in the $L_0$-evasions (darker means feature is more often picked).}
\vspace{-1.5em}
\end{figure}

\subsection{Robustness}
 For each learned model, and for all of 
the 100 correctly classified evaluation instance, we compute the optimal (or best effort) solution to the evasion problem 
under all of the deformation metrics. We use the Gurobi~\cite{gurobi} solver to compute the optimal
evasions for all distances and all models but NN and RBF-SVM. We use a classic projected gradient descent method for solving the $L_1, L_2$ 
and $L_\infty$ evasions of NN and RBF-SVM, and address the $L_0$-evasion case by an iterative coordinate 
descent algorithm and a brute force grid search at each iteration. 
Figure~\ref{fig:comp} summarizes the obtained adversarial bounds as one boxplot for each combination of model and distance. 
Although the tree ensembles BDT and RF have very competitive accuracies, they systematically rank at the bottom 
for robustness across all metrics. Remarkably, negligible $L_1$ or $L_2$ perturbations suffice to evade those models. RBF-SVM is apparently the hardest model to evade, agreeing with 
results from~\cite{goodfellow2014}. 
NN and CPM exhibit very similar performance despite having quite different architectures. 
Finally, the $L_1$-regularized linear model exhibits significantly more brittleness than its $L_2$ counterpart.
This phenomenon is explained by large weights concentrating in specific dimensions as a result of sparsity. 
Thus, small modifications in the heavily weighted model dimensions can result in large classifier output variations.

\subsection{Hardening by Adversarial Boosting}
\label{subsec:hardening}
We empirically demonstrate how to significantly improve the robustness of the BDT model 
by adding evading instances to the training set during the boosting process. At each boosting round, we use our fast 
\emph{symbolic prediction}-based algorithm to create budgeted ``adversarial'' instances with respect to the current model 
and for all the 11,876 original training instances. For a given training instance $x$ with label $y$ 
and a modification budget $B \geq 1$, a budgeted ``adversarial'' training instance $x^*$ is such 
that $\|x - x^*\|_0 \leq B$ and the margin $yf(x^*)$ is as small as possible. Here, we use $B=28$, the size of 
the picture diagonal, as our budget. The reason is that modifying 28 pixels over 784 is not enough 
to perceptually morph a handwritten ``2'' into ``6''. 
The training dataset for the current round is then formed by appending to the original training dataset 
these evading instances along with their correct labels, thus increasing the size of the training set by a factor 2. 
Finally, gradient boosting produces the next regression tree which by definition minimizes the error of the 
augmented ensemble model on the adversarially-enriched training set. After 1,000 adversarial boosting rounds, our model
has encountered more than 11 million adversarial instances, without ever training on more than 24,000 instances at a time.

We found that we needed to increase the maximum tree depth from 4 to 6 in order to obtain an acceptable error rate. 
After 1,000 iterations, the resulting model BDT-R has a slightly higher testing accuracy than BDT (see Table~\ref{table:models}). 
Unlike BDT, BDT-R is extremely challenging to optimally evade using the MILP solver:
the branch-and-bound search continues to expand nodes after 1 day on a 6 core Xeon 3.2GHz machine. To obtain the tightest 
possible evasion bound, we warm-start the solver with the solution found by the fast evasion technique and report the best
solution found by the solver after an hour. Figure~\ref{fig:comp} shows that BDT-R is more robust than our previous 
champion RBF-SVM with respect to $L_0$ deformations. Unfortunately, we found significantly lower scores
on all $L_1, L_2$ and $L_\infty$ distances compared to the original BDT model: 
hardening against $L_0$-evasions made the model more sensitive to all other types of evasions.

\label{sub:retraining}

\section{Conclusion}
We have presented two novel algorithms, one exact and one approximate, for systematically computing 
evasions of tree ensembles such as boosted 
trees and random forests. On a classic digit recognition task, both gradient boosted trees and random forests 
are extremely susceptible to evasions. We also introduce \emph{adversarial boosting} and show that it trains 
models that are hard to evade, without sacrificing accuracy. One future work direction would be to use
these algorithms to generate ``small'' evading instances for practical security systems. Another direction would
be to better understand the properties of adversarial boosting. In particular, it is not known whether this hardening
approach would succeed on all possible datasets.

\newpage
\section*{Acknowledgements}
This research is supported in part by Intel's ISTC for Secure
Computing, NSF grants 0424422 (TRUST) and 1139158,
the Freedom 2 Connect Foundation, US State Dept. DRL,
LBNL Award 7076018, DARPA XData Award FA8750-12-2-0331, 
and gifts from Amazon, Google, SAP, Apple, Cisco,
Clearstory Data, Cloudera, Ericsson, Facebook, GameOnTalis,
General Electric, Hortonworks, Huawei, Intel, Microsoft,
NetApp, Oracle, Samsung, Splunk, VMware, WANdisco
and Yahoo!. The opinions in this paper are those of the
authors and do not necessarily reflect those of any funding
sponsor or the United States Government.
 

\bibliography{paper}
\bibliographystyle{icml2016}

\end{document}